\documentclass{article}
\usepackage{ijcai23}

\usepackage{times}
\usepackage{soul}
\usepackage{url}
\usepackage[hidelinks]{hyperref}
\usepackage[utf8]{inputenc}
\usepackage[small]{caption}
\usepackage{graphicx}
\usepackage{amsmath}
\usepackage{amsthm}
\usepackage{booktabs}
\usepackage{algorithm}
\usepackage{algorithmic}
\usepackage[switch]{lineno}
\usepackage{multirow}
\usepackage{makecell}
\usepackage{newfloat}
\usepackage{listings}
\usepackage{amsmath,amsfonts}

\title{Selecting Learnable Training Samples is All DETRs Need in Crowded Pedestrian Detection}

\author{
    Author Name
    \affiliations
    Affiliation
    \emails
    email@example.com
}

\author{
Feng Gao$^1$
\and
Jiaxu Leng$^1$\and
Ji Gan$^1$\And
Xinbo Gao$^1\dag$
\affiliations
$^1$Chongqing University of Posts and Telecommunications\\
\emails
\{d210201005\}@stu.cqupt.edu.cn,
\{lengjx, ganji, gaoxb\}@cqupt.edu.cn
}
\begin{document}
\maketitle
\begin{abstract}
DEtection TRansformer (DETR) and its variants (DETRs) achieved impressive performance in general object detection. However, in crowded pedestrian detection, the performance of DETRs is still unsatisfactory due to the inappropriate sample selection method which results in more false positives. To settle the issue, we propose a simple but effective sample selection method for DETRs, Sample Selection for Crowded Pedestrians (SSCP), which consists of the constraint-guided label assignment scheme (CGLA) and the utilizability-aware focal loss (UAFL). Our core idea is to select learnable samples for DETRs and adaptively regulate the loss weights of samples based on their utilizability. Specifically, in CGLA, we proposed a new cost function to ensure that only learnable positive training samples are retained and the rest are negative training samples. Further, considering the utilizability of samples, we designed UAFL to adaptively assign different loss weights to learnable positive samples depending on their gradient ratio and IoU. Experimental results show that the proposed SSCP effectively improves the baselines without introducing any overhead in inference. Especially, Iter Deformable DETR is improved to 39.7(-2.0)\% MR on Crowdhuman and 31.8(-0.4)\% MR on Citypersons.
\end{abstract}
\section{Introduction}
Pedestrian detection is a critical yet challenging research field in computer vision. It is widely applied in autonomous driving and surveillance. In the last decade, CNN-based detectors have dominated the development of pedestrian detection, which has hugely succeeded \cite{4,5,7,2018,8,11,12,14}. The performance of pedestrian detection on the Reasonable subset of Caltech dataset \cite{Caltech} is nearly saturated, with 2.31\% MR by PedHunter \cite{pedhunter}.

\begin{figure}[!t]
\centering
\includegraphics[width=3.4in]{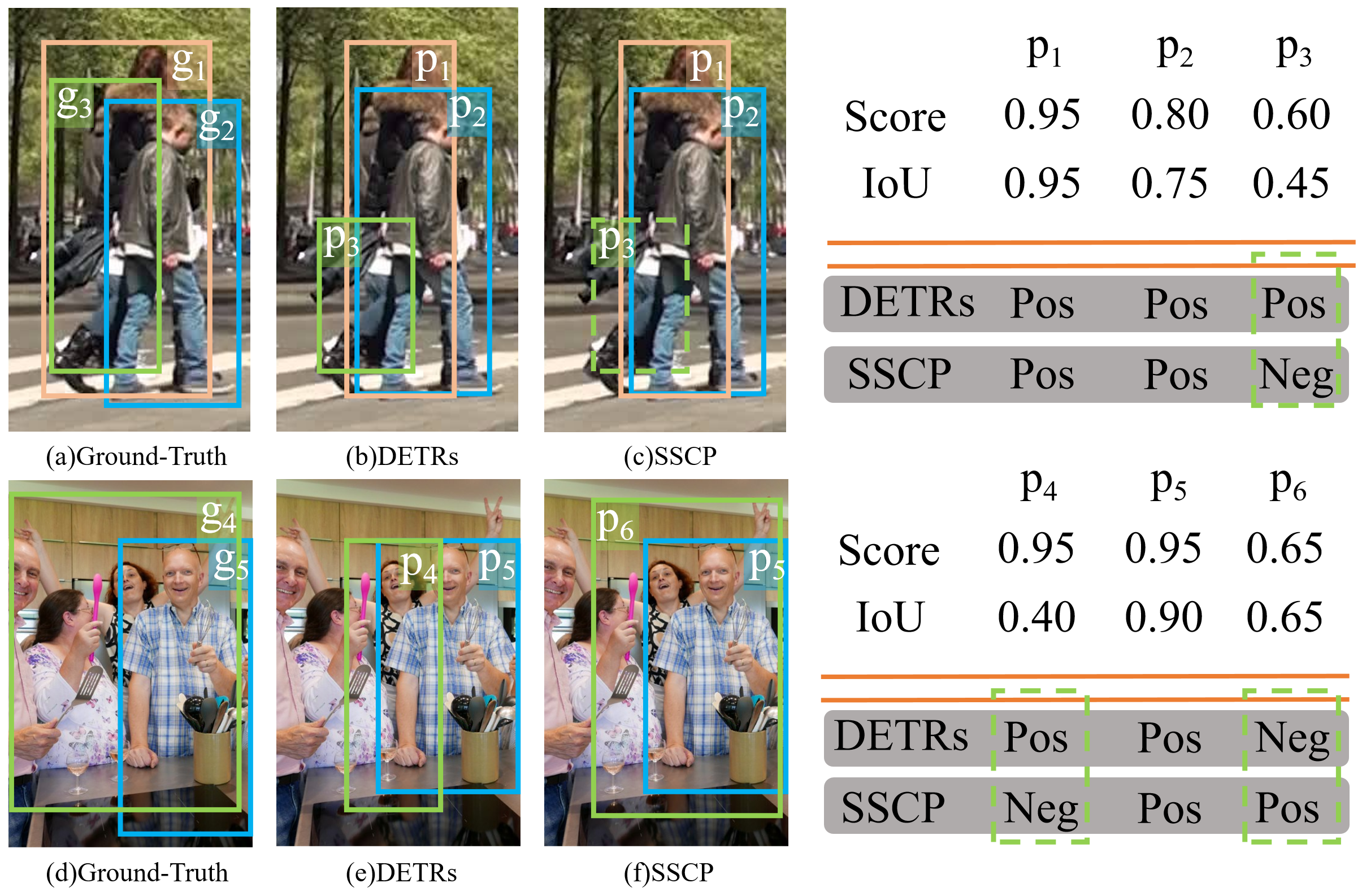}
\caption{The sample selection method of DETRs may select positive training samples which are not learnable. $g_i$ are ground-truth boxes. $p_j$ are predictions. The solid box means it is selected as a positive training sample. The dashed box means it is selected as a negative training sample. Our SSCP selects learnable positive training samples and filters out positive training samples which are not learnable and converts them into negative training samples.}
\label{fig1}
\end{figure}

\par However, the performance of pedestrian detection drops a lot in crowded scenes. Taking PedHunter as an example, its MR is 8.32\% when it is applied to the Reasonable subset of Citypersons \cite{Citypersons}. However, its MR is 43.53\% on the Heavy Occlusion subset of Citypersons.
\par So far, almost all pedestrian detectors are CNN-based. The main challenge comes from a component of CNN-based pedestrian detectors, non-maximum suppression (NMS). Since CNN-based pedestrian detectors produce multiple predictions for each pedestrian,  CNN-based pedestrian detectors need NMS to remove redundant predictions. In crowded scenes, correct predictions are removed inevitably by NMS due to heavy overlaps. For example, on Crowdhuman dataset \cite{Crowdhuman}, Sun et. al \cite{onenet} applied NMS on annotation boxes only obtains 95\% recall. Hence, CNN-based pedestrian detectors with NMS are difficult to achieve the ideal results.

\par Recently, a novel object detection framework, DETR \cite{DETR} was proposed. Subsequently, its variations achieved dominant performance in general object detection \cite{DDETR,srcnn}. DETRs are ideal pedestrian detectors in theory because they are NMS-free. These works inspired some pioneers \cite{PETR,IDDETR} to research DETRs for crowded pedestrian detection.
The two studies found that compared with Faster-R-CNN-FPN \cite{FPN}, Deformable DETR \cite{DDETR} produces more false positives in crowded pedestrian detection.
\iffalse They both argue the defect is that the attention module in decoder layers cannot distinguish well between the background and pedestrians in crowded scenes.\fi

\par \textbf{ We argue that DETRs produce more false positives in crowded pedestrian detection mainly due to the sample selection method of DETRs}. Figure 1 shows two examples. In the first example, as shown in Figure 1(a)(b), pedestrian $g_3$ is heavily occluded by pedestrians $g_1$ and $g_2$. Due to DETRs' detection mechanism, the certain decoder layer may not produce a learnable positive training sample for pedestrian $g_3$. In this case, DETRs still select $p_3$ as the positive training sample for $g_3$ \iffalse which is with low IoU but high confidence\fi. We argue that $p_3$ is not learnable because $p_3$ lacks effective features. Using $p_3$ as a positive training sample leads DETRs to detect the background as pedestrians. It is a kind of false positive. In the second example, as shown in Figure 1(c)(d), DETRs select $p_4$ as the positive training sample for $g_4$. Although $p_4$ obviously contains the effective features of $g_4$, we still argue that $p_4$ is not learnable because the extremely large location error confuses the training. Using $p_4$ as a positive training sample leads DETRs to produce false positives due to the inaccurate location.

\par To solve the problem mentioned above, we propose a simple but effective sample selection method, Sample Selection for Crowded Pedestrians (SSCP). The proposed SSCP consists of two parts, the constraint-guided label assignment scheme (CGLA) and the utilizability-aware focal loss (UAFL). In CGLA, we design a new cost function. This cost function contains two constraints. First, the cost between each sample and ground-truth pair is calculated. Then the Hungarian algorithm is used to select positive and negative training samples based on the cost. Finally, CGLA selects learnable positive training samples for training and filters out the positive training samples that do not satisfy the constraints and convert them into negative training samples, as shown in Figure 1(c)(f). UAFL turns the fixed label $y$ and $\gamma$ in Focal Loss \cite{FL} into adaptive variables. In UAFL, the label $y$ changes with the IoU of the sample. $\gamma$ is related to the IoU and the gradient ratio.  More importantly, SSCP can be plugged into any DETRs and does not participate in inference, so it improves DETRs without any overhead. Extensive experiments on Crowdhuman and Citypersons datasets support our analysis and conclusions. With the proposed SSCP, the state-of-the-art performance is improved to 39.7\% MR on Crowdhuman and 31.8\% MR on Citypersons.
\section{Related Work}
\subsection{Crowded Pedestrian Detection}
\par Faster-R-CNN-FPN is the broadest baseline in pedestrian detection. N2NMS \cite{N2NMS} improves Faster-R-CNN-FPN to simultaneously detect the visible part and the full body of a pedestrian and use the less overlapped visible part for NMS. MIP \cite{MIP} improves Faster-R-CNN-FPN to detect two objects by one proposal and does not apply NMS to the objects which are from the same proposal. 

\par DETR is a kind of NMS-free detector. However, due to the fact that DETR originated in 2020, only a few pioneers have researched pedestrian detection based on DETR. They \cite{PETR,IDDETR} found DETR produces more false positives than CNN-based pedestrian detectors in crowded scenes. They believed the fundamental problem lies in the inability of the cross-attention module to distinguish well between the pedestrian and the background. Liu et al.\cite{PETR} proposed a novel cross-attention module that utilized visible parts to enhance the feature extraction of crowded pedestrians. Zheng et al.\cite{IDDETR} designed a relation network to model the relationship among detected pedestrians. The relation network utilized high-confidence pedestrians to detect near low-confidence pedestrians. The two pieces of research greatly inspired us to research DETRs' potential in crowded pedestrian detection. 
\subsection{Label Assignment}
\par Faster-R-CNN \cite{fasterrcnn} is the most representative anchor-based two-stage object detector. It selects positive training samples based on IoU between each anchor box and ground-truth pair.
FCOS \cite{fcos} defines the anchor points near the center of ground-truths as positive training samples. 
CenterNet \cite{Centernet} defines the anchor point which is the closest to the center point of the ground-truth as the positive training sample.
Recently, specialized studies about label assignment began to emerge, since the importance of label assignment is realized. Zhang et al. \cite{atss} proposed an adaptive label assignment scheme, ATSS, to select positive training samples based on the statistical characteristics of predictions. With ATSS, the performance gap between RetinaNet \cite{FL} and FCOS has narrowed significantly. Ge et al. \cite{ota} regarded label assignment as optimal transport assignment, which selects positive training samples by Sinkhorn-Knopp Iteration \cite{Sinkhorn}. 
Label assignment of DETR is different from CNN-based detectors, which not only considers position such as IoU but also classification. The cost between each sample and ground-truth pair is calculated and DETR selects training samples based on the cost by the Hungarian algorithm. \iffalse Sun et al. \cite{onenet} transplanted DETR's label assignment scheme to FCOS, which turns FCOS into an NMS-free object detector.\fi

\subsection{Loss Function}
\par Loss function plays an important role in computer vision. In pedestrian detection, a loss function commonly consists of two parts: regression loss function and classification loss function. Commonly used regression loss functions include SmoothL1 loss \cite{l1}, IoU loss \cite{iou}, GIoU loss \cite{giou}, and the most commonly used classification loss function is Focal Loss. 
Adding extra items or designing a new loss function is a common method to improve crowded pedestrian detection. Wang et al.\cite{repulsion} proposed repulsion loss to promote the predicted bounding box farther away from other predicted bounding boxes and other ground-truths. Tang et al. \cite{autopedestrian} proposed a search scheme to automatically search parameters for loss function in pedestrian detection.

\section{Method}
In this section, we briefly analyse the pipeline of DETRs and point out the drawbacks in crowded pedestrian detection. And then, we elaborate our sample selection method for DETRs in crowded pedestrian detection, Sample Selection for Crowded Pedestrians (SSCP), which consists of the constraint-guided label assignment scheme (CGLA) and the utilizability-aware focal loss(UAFL).
\subsection{Pipeline of Deformable DETR}
To illustrate the proposed method, we choose Deformable DETR as an example to introduce the pipeline of DETRs and analyze its defect in crowded pedestrian detection. Its pipeline can be formulated as:

\begin{equation}
\begin{aligned}
   x^{fpn}& \gets Backbone(img)  \\
 x^{enc} & \gets Enc(x^{fpn}) \\
  ref_{0},e^{q}& \gets Split(query)  \\
p_t,c_t,x^{dec}_t,ref_{t} & \gets Dec_{t-1}(x^{ecn},ref_{t-1},e^{q},x^{dec}_{t-1}) \\
 cost_t &\gets Cost(p_{t},gt) \\ 
index_t &\gets H(cost_t) \\ 
loss_t&\gets Loss(p_{t},gt,index_t). 
\end{aligned}
\end{equation}

\par An image is input into a backbone to produce muti-scale feature maps $x^{fpn}$. The feature maps $x^{fpn}$ are input into transformer encoder layers $Enc$ to produce feature map $x^{enc}$. $query$ denotes the trainable object query. $query$ is split into two parts, reference points $ref_0$ and query embedding $e^q$. The reference point is like the anchor point or anchor box. $Dec$ denotes the decoder layer. In decoder layers, the query embedding $e^q$ plus $x^{dec}$ are used to predict the offsets. The reference points $ref_{t-1}$ extract features from $x^{ecn}$ based on the offsets. The extracted features is $x^{dec}_{t}$. $x^{dec}_{t}$ is input into detection heads to produce predictions $p_t$. The predicted bounding boxes of prediction $p_t$ and $ref_{t}$ are the same things. $ref_{t}$ is used for the next decoder layer for refinement. Each prediction of $p_t$ is a sample. Each sample of $p_t$ and ground-truth $gt$ pair is computed to produce a cost matrix $cost_t$. The cost formulation is as follows:
\begin{equation}
Cost=\lambda _{1}(C_{cls}+C_{L1})-\lambda _{2}C_{GIoU}.
\end{equation}

The Hungarian algorithm $H$ selects positive and negative training samples based on the cost matrix $cost_t$, as shown in the left of Figure 2. To diagrammatize clearly, we only draw the center points of $p_t$ instead of the bounding boxes. Finally, the loss of each training sample and ground-truth pair is computed.

\begin{figure}[!t]
\centering
\includegraphics[width=3.0in]{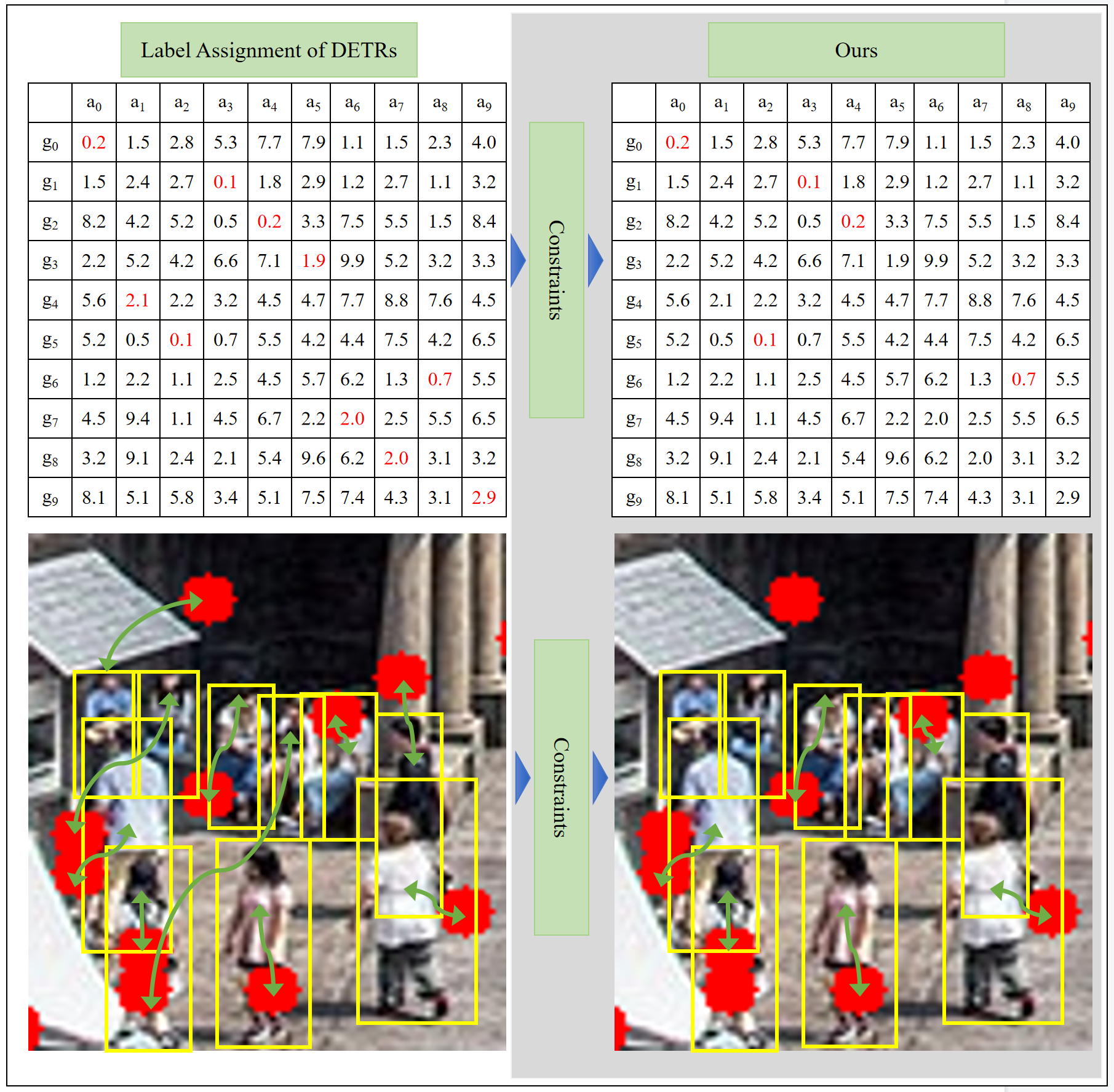}
\caption{Label assignment visualization of DETRs and CGLA. The left is label assignment of DETRs. The whole is our proposed CGLA.}
\label{la}
\end{figure}

\begin{figure}[!t]
\centering
\includegraphics[width=3.0in]{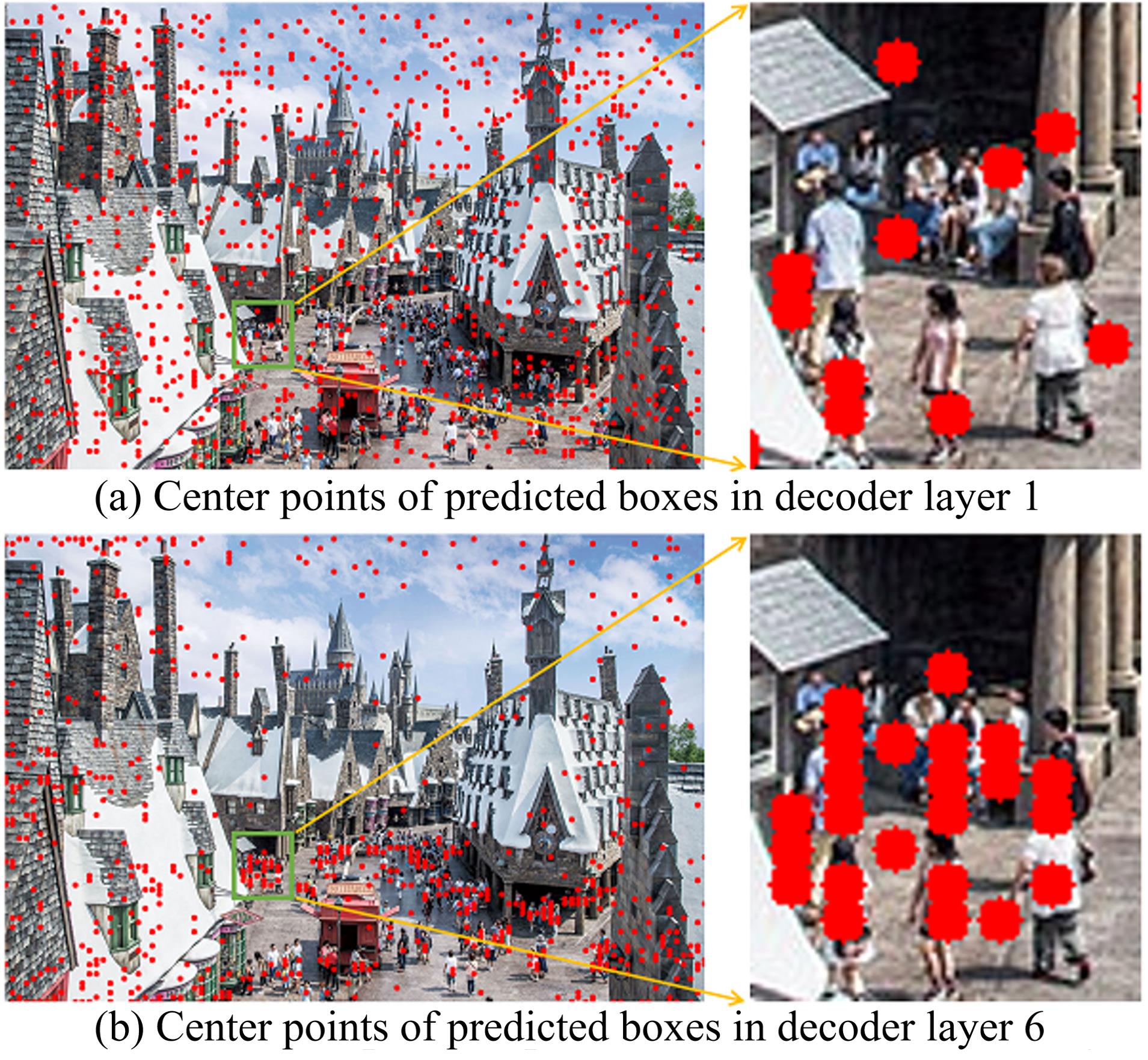}
\caption{Visualization of center points of predicted boxes in decoder layers.}
\label{fig2}
\end{figure}

\subsection{Analysis}
In decoder layers, predicted bounding boxes of $p_t$ gradually converge to pedestrians. As shown in Figure \ref{fig2}, predicted bounding boxes of $p_t$ may be not able to cover all pedestrians, because of severe occlusion and crowding, especially in the forward decoder layers such as decoder layer 1. The Hungarian algorithm is a one-to-one assignment algorithm. Each ground-truth must be assigned with a sample even though the sample doesn't contain effective features or causes extremely large regression loss, as shown in the left of Figure \ref{la}. These positive training samples promote DETRs to produce more false positives in crowded scenes. Therefore, these samples are not learnable. There are only two kinds of samples in pedestrian detection, pedestrians or the background. Therefore, we argue that these positive training samples which are not learnable are negative training samples.

\subsection{Constraint-guided Label Assignment Scheme}
\par To select learnable positive training samples and filter out positive training samples which are not learnable and convert them into negative training samples, we propose the constraint-guided label assignment scheme (CGLA). We argue that a learnable positive training sample should contain corresponding effective features and should not cause extreme regression loss. Therefore, a learnable positive sample should overlap with the corresponding ground-truth in some degree. We design two constraints, the center constraints and the position constraint. If the positive training sample satisfies the two constraints, we think it is learnable. Otherwise, the positive training sample will be converted into a negative training sample, as shown in Figure \ref{la}. The details are described below.

\par Firstly, we design a novel cost for the Hungarian algorithm. In DETRs, the original cost consists of three parts: classification cost, GIoU cost, and L1 cost. L1 cost and GIoU cost have a similar function, as mentioned in Equation (2). However, L1 cost is difficult to determine whether the sample contains the corresponding features. For example, with the same GIoU cost, L1 cost of the larger scale sample is more than the smaller scale sample. Therefore, we replace the L1 cost with the center constraints cost and the position constraint cost. The formulation is as follows:\begin{equation}
Cost=\lambda _{1}C_{cls}-\lambda _{2}(C_{GIoU}+C_{pos}+C_{cenx}+C_{ceny}),
\end{equation}
\begin{figure}[!t]
\centering
\includegraphics[width=3.0in]{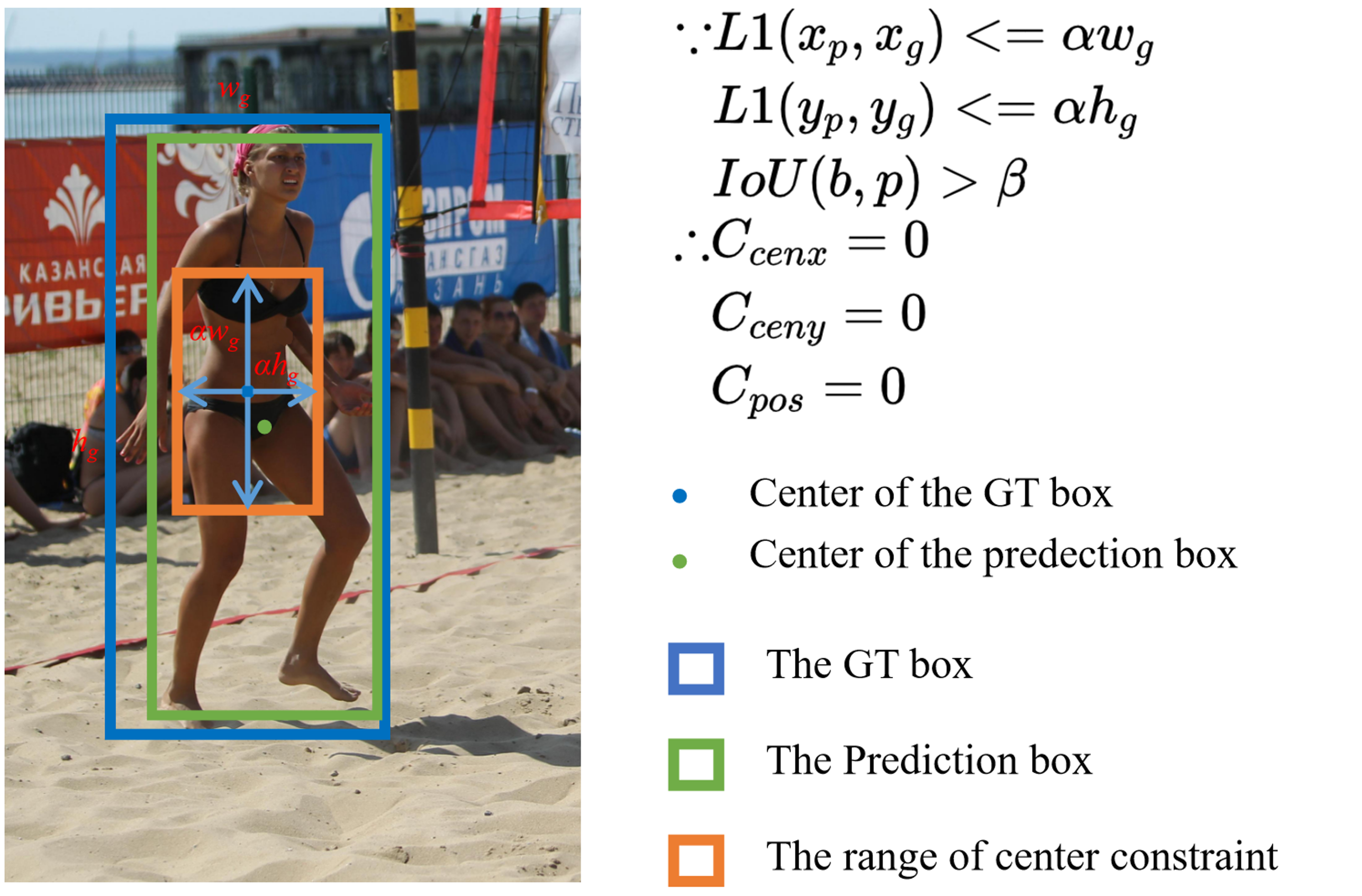}
\caption{An example of the proposed center and position constraints.}
\label{fig4}
\end{figure}
where $C_{cenx}$, $C_{ceny}$ and $C_{pos}$ denote two center constraints cost and the position constraint cost. According to the feature extraction mechanism of DETRs mentioned in 3.1, we argue that if a sample contains sufficient effective features, the distance between the center of the sample and the corresponding ground-truth center should be within a range. Therefore, we design the center constraints. The specific formulation is as follows:
\begin{equation}
C_{cenx}^{ij}=\left\{\begin{matrix} 
  -1, L1(x_{p}^{i},x_{g}^{j})>\alpha w_{g}^{j}\\  

  0, L1(x_{p}^{i},x_{g}^{j})<=\alpha w_{g}^{j},\\ 

\end{matrix}\right. 
\end{equation}
\begin{equation}
C_{ceny}^{ij}=\left\{\begin{matrix} 

  -1, L1(y_{p}^{i},y_{g}^{j})>\alpha h_{g}^{j}\\

  0, L1(y_{p}^{i},y_{g}^{j})<=\alpha h_{g}^{j},\\
\end{matrix}\right. 
\end{equation}
where $\alpha$ denotes the parameter of the center constraints cost. $x_{p}^{i}$ and $x_{g}^{j}$ denote the center coordinates of a sample $i$ and a ground-truth $j$ on x-axis. $L1$ denotes L1 distance between $x_{p}^{i}$ and $x_{g}^{j}$. $w_{g}^{j}$ denotes the width of ground truth $j$. Equation (5) is similar to Equation (4).

\par The IoU between the sample and the ground-truth not only reflects the number of effective features but also the regression loss. Therefore, we design the position constraint. The formulation is as follows:

\begin{equation}
C_{pos}^{ij}=\left\{\begin{matrix} 

  -1, IoU(p^{i},g^{j})<=\beta \\

  0, IoU(p^{i},g^{j})>\beta, \\
\end{matrix}\right. 
\end{equation}
where $\beta$ denotes the IoU threshold of the position constraint cost. $p^{i}$ and $g^{j}$ denote bounding boxes of a sample $i$ and a ground truth $j$, respectively.

\par  In CGLA, the cost between each sample and ground-truth pair is calculated to produce the cost matrix based on our new cost function, as shown in Equation (3). Based on the cost matrix, the Hungarian algorithm assigns each ground-truth to a sample. The next step of our algorithm is to filter out positive training samples which are not learnable by the constraints. Figure 4 is an example to show a positive training sample, which satisfies all constraints. The positive training samples which satisfy the constraints are learnable positive training samples. Otherwise, they are not learnable and are converted to negative training samples. Algorithm 1 describes how CGLA works.

\begin{algorithm}[tb]
\caption{Constraint-guided label assignment scheme}
\label{alg:algorithm}
\textbf{Input}:
\par \qquad $ \textbf{G is a set of ground truths} $
\par \qquad $ \textbf{P is a set of samples} $
\par \qquad $ \textbf{$\alpha$ is a parameter of the center constraint cost}  $
\par \qquad $ \textbf{$\beta$ is a parameter of the position constraint cost}  $

\textbf{Output}: 
\par \qquad $ \textbf{Pos is a set of positive training samples } $
\par \qquad $ \textbf{Neg is a set of negative training samples } $

\begin{algorithmic}[1] %[1] enables line numbers
\STATE pairwise center constraint cost on x axis:\par
\hspace{0.275cm}$c^{ij}_{cenx}$=L1($P^{reg}_{i}$,$G^{reg}_{j}$,$\alpha$)
\STATE pairwise center constraint cost on y axis:\par
\hspace{0.275cm}$c^{ij}_{ceny}$=L1($P^{reg}_{i}$,$G^{reg}_{j}$,$\alpha$)

\STATE pairwise position constraint cost:\par
\hspace{0.275cm}$c^{ij}_{pos}$=IoU($P^{reg}_{i}$,$G^{reg}_{j}$,$\beta$)
\STATE pairwise classification cost:$c^{ij}_{cls}$=FocalLoss($P^{cls}_{i}$,$G^{cls}_{j}$)
\STATE pairwise GIoU cost:$c^{ij}_{giou}$=GIoULoss($P^{reg}_{i}$,$G^{reg}_{j}$)
\STATE label assignment based on Hungarian Algorithm:\par \hspace{0.275cm}Pos,Neg=H($c_{giou}$,$c_{cls}$,$c_{cenx}$,$c_{ceny}$,$c_{pos}$)
\STATE filter positive samples which are not learnable out from Pos:
\par \hspace{0.275cm}for m in Pos:
\par \hspace{0.5cm}if $c^{m}_{cenx}$==-1 or $c^{m}_{ceny}$==-1 or $c^{m}_{pos}$==-1:
\par \hspace{1cm}delete m from Pos and add m into Neg
\STATE \textbf{return} Pos, Neg
\end{algorithmic}
\end{algorithm}

\subsection{Utilizability-aware Focal Loss}
CGLA filters out positive training samples which are not learnable. However, the utilizability of positive training samples is different. The positive training sample with a larger IoU is easier to learn. That the positive training sample makes a higher gradient ratio means the sample is not learned well. The gradient ratio donates the ratio of the gradient produced by a sample as a positive and negative training sample in the loss function. We combine the two factors to represent the utilizability of the positive training samples and improve Focal Loss based on the utilizability. 
\par In Focal Loss, each sample corresponds to a fixed label $y$ and a fixed parameter $\gamma$. A typical Focal Loss is as follows(we ignore $\alpha$ in the original paper for simplicity):
\begin{equation}
L_{FL}=-(1-p)^\gamma ylogp-p^\gamma (1-y)log(1-p).
\end{equation}
Specifically, the positive label $y=1$ in Focal Loss is changed to \textbf{the soft label} $0<y<1$ in UAFL. The value of the soft label $y$ is the IoU between the positive training sample and the corresponding gound-truth. The core idea is not to force a sample with a smaller IoU to predict a high confidence, because it contains fewer effective features.\textbf{ We convert $\gamma$ to an adaptive parameter}. Each positive training sample corresponds to a $\gamma_j$ whose value adaptively changes with its IoU and gradient ratio. The formula of $\gamma$ is:

\begin{equation}
\begin{aligned}
\gamma= &\gamma_o + \gamma_g^i\\
      = &\gamma_o + (g^i-t_g)(\beta-y),\\
\end{aligned}
\end{equation}
where $\gamma_o$ is a fixed parameter that is like an anchor. The parameter $g^i$ indicates the gradient ratio of the $i$th training sample. The $t_g$ is a self-adaptive threshold which is the mean value of $g$. The $\beta$ indicates a threshold that is as same as $\beta$ in CGLA. 
To insure the stability of training, we clamp the value of $g^i-t_g$ in the range [0,3]. The value of $g^i-t_g$ reflects if the sample is learned well. For example, the larger value of $g^i-t_g$ means the sample is learned worse. The value of $\beta-y$ reflects if the sample is worth utilizing. For example, the smaller value of $\beta-y$ means the sample is more worth utilizing. Combining the two factors, UAFL can adaptively regulate loss weights.
The final formula of UAFL is:

\begin{equation}
L_{UAFL}=-|y-p|^\gamma (ylogp+(1-y)log(1-p)),
\end{equation}

When the confidence of the prediction equals its IoU, the loss is optimal. 

\section{Experiments}
%\par In this section, we conduct our method on two challenging pedestrian detection datasets, Crowdhuman and Citypersons, to demonstrate the generality and effectiveness of the proposed methods in different scenes. 
\subsection{Experimental Settings}
\par \textbf{Datasets.} we conduct our method on two challenging pedestrian detection datasets, Crowdhuman and Citypersons, to demonstrate the generality and effectiveness of the proposed methods in different scenes.
\par Crowdhuman is a heavily crowded pedestrian detection dataset published by Megvii with 15000 images in the training set, 4370 images in the validation set and 5000 images in the test set.  In Crowdhuman, on average, each image contains 22.6 pedestrians. Pairwise pedestrians with IoU over 0.3 is about 9 pairs per image, and triples of pedestrians with IoU over 0.3 is about 0.5 pairs per image.  
\par Citypersons is a recently released challenging pedestrian detection dataset with 2975 images in the training set, 500 images in the validation set and 1575 images in the test set.  The bounding box annotations of the full body and visible part are provided. In Citypersons, about 70\% of the pedestrian instances are occluded in different degrees, so it is also quite suitable for verifying the effectiveness of pedestrian detectors in crowded scenes. 
\par \textbf{Evaluation metrics.} Log-average miss rate(MR) is most the commonly used evaluation metric in pedestrian detection. MR evaluates the comprehensive performance of the detector in high-confidence intervals. The smaller MR indicates a better performance.
\par \textbf{Implementation Details.}  The proposed method is implemented in PyTorch and trained on a RTX 3090 GPU. Most training details are the same with Iter Deformable DETR \cite{IDDETR}. The differences include: (1) At the stage of fine-tuning, we introduce CGLA and UAFL for 20 epochs because at the beginning of training, few positive training samples satisfy the constraints, which leads to training instability. (2) For Citypersons, the training and testing size are 1024$\times$2048 and we only use horizontal flip as the data augmentation. 

\subsection{Comparison with State-of-the-Arts}
For Crowdhuman, we compare our method with two baselines, Deformable DETR and Iter Deformable DETR. The former is a general object detector. The latter is a crowded pedestrian detector. Deformable DETR is the baseline of Iter Deformable DETR. With our methods, Deformable DETR achieves similar performance to Iter Deformable DETR without introducing any overhead in inference. We make comparisons with the state-with-the-arts: Faster-R-CNN-FPN \cite{FPN}, MIP \cite{MIP}, R2NMS \cite{N2NMS}, AEVB \cite{vpd}, AutoPedestrian \cite{autopedestrian}, OAF-Net \cite{OAF} and DMSFLN \cite{DMSFL}. In Table 1, our method achieves state-of-the-art performance, 39.7\% MR, which fully outperforms the best Iter Deformable DETR by 1.8\%.

\begin{table}[!t]
\caption{Comparison with the state-of-the-arts on Crowdhuman. \label{tab:table}}
\centering
\begin{tabular}{cccc}
\toprule[1.4pt]
Model &Backbone &MR$\downarrow$\\
\hline
FPN (CVPR'17)& ResNet-50&42.9 \\

MIP (CVPR'20) &ResNet-50& 41.4 \\

R2NMS (CVPR'20) &ResNet-50 &43.4 \\

AEVB (CVPR'21) &ResNet-50 &40.7 \\

AutoPedestrian (TIP'21)&ResNet-50 &40.6\\

DMSFLN(TITS'21)&VGG-16 &43.6 \\

OAF-Net(TITS'22)&HRNet-w32 &45.0 \\
\hline
D-DETR (ICLR'21)&ResNet-50 &44.6 \\
D-DETR+Ours&ResNet-50 &42.0   \\
\hline
Iter-D-DETR (CVPR'22) &ResNet-50 &41.5 \\
Iter-D-DETR (our implementation) &ResNet-50 &41.9 \\
Iter-D-DETR+Ours&ResNet-50 &\textbf{39.7} \\
\toprule[1.4pt]
\end{tabular}
\end{table}

\iffalse
\begin{figure*}[!t]
\centering
\includegraphics[width=7.0in]{CGLAAB.png}
\caption{Performance under different parameters on Crowdhuman. \label{tab:table}}
\label{fig6}
\end{figure*}
\fi

\begin{table}[!t]
\caption{Comparison with the state-of-the-arts on Citypersons.The superscript $\dag$ indicates the pedestrians over 50 pixels in height with more than 35\% occlusion, instead of pedestrians over 50 pixels in height with 35-80\% occlusion. Thus, $\dag$ suggests higher difficulty. \label{tab:table}}
\centering
\begin{tabular}{cccccc}
\toprule[1.4pt]
Model &Backbone &R$\downarrow$ & HO$\downarrow$\\
\hline

AP2M (AAAI'21)& ResNet-50&10.4 & 48.6\dag\\
CoupleNet(TITS'21)& ResNet-50&12.4 & 49.8\dag\\

FAPD(TITS'22)& VGG-16&10.9 & 47.5\dag\\

DPFD(TITS'22)& VGG-16&10.3 & 50.2\dag\\

FC-Net(TITS'22)& ResNet-50&13.5 & 44.3\dag\\

OAF-Net(TITS'22)& ResNet-50&9.4 & 43.1\dag\\
\hline
Iter-D-DETR (CVPR'22)& ResNet-50&10.4 & 40.8\dag\\
Iter-D-DETR+CGLA & ResNet-50&10.5 & 40.6\dag\\
Iter-D-DETR+SSCP & ResNet-50&10.4 & \textbf{40.0}\dag\\
\hline \hline
MGAN+(TIP'21)& VGG-16&11.0 & 39.7\\

KGSNet(TNNLS'21)& VGG-16&11.0 & 39.7\\

DMSFLN(TITS'21)& VGG-16&9.9 & 38.1\\
\hline
Iter-D-DETR (CVPR'22)& ResNet-50&10.4 & 32.2\\
Iter-D-DETR+CGLA & ResNet-50&10.5 & 32.0\\
Iter-D-DETR+SSCP & ResNet-50&10.4 & \textbf{31.8}\\
\toprule[1.4pt]
\end{tabular}

\end{table}

\par For Citypersons, We compare our methods with several state-of-the-art pedestrian detectors on Reasonable(R) subset, Heavy Occlusion(HO) subset and Heavy Occlusion(HO)$\dag$ subset, including AP2M \cite{AP2M}, CounpleNet \cite{Coupled}, FAPD \cite{FAPD}, DPFD \cite{DPFD}, FC-Net \cite{fc}, OAF-Net, MGAN+ \cite{MGAN+} ,KGSNet \cite{KGSnet} and DMSFLN. The pedestrians are all over 50 pixels in height. R indicates the pedestrians are with occlusion less than 35\%. HO$\dag$ indicates the pedestrians are with occlusion more than 35\%. HO indicates the pedestrians are with occlusion between 35\% to 80\%. In Table 2, our method achieves state-of-the-art performance, 40.0\% MR on HO$\dag$ subset and 31.8\% MR on HO subset, which outperforms the best Iter Deformable DETR by 0.8\% MR on HO$\dag$ subset and 0.4\% MR on HO subset. \iffalse For fairness, the testing size is 1024$\times$2048 except OAF-Net, because its testing size of the best performance is 640$\times$1280.\fi

\subsection{Ablation studies}
We conduct ablation studies on Crowdhuman and use Iter Deformable DETR as our baseline. As shown in Table 3, our implemented Iter Deformable DETR achieves 41.9\% MR on Crowdhuman. To evaluate the effectiveness of SSCP, we first conducted the center constraints alone, which improves MR by 1.3\%. We further add the position constraint and MR reaches 40.0\% MR. Finally, We further introduce UAFL and achieve 39.7\% MR.
\begin{table}[!t]
\caption{Contributions of each component on Crowdhuman. The CC indicates the center constraints cost. The PC indicates the position constraint cost. The SL indicates the soft label $y$. The AG indicates the adaptative $\gamma$. \label{tab:table}}
\centering
\begin{tabular}{c|cccc|c}
\toprule[1.4pt]
\multirow{2}*{Method}&  \multicolumn{2}{c}{CGLA} & \multicolumn{2}{c|}{UAFL}& \multirow{2}*{MR$\downarrow$} \\
&CC  & PC  & SL& AG&   \\
\hline
\multirow{7}*{\makecell[c]{Iter-D-DETR \\ \cite{IDDETR}}}&& & && 41.9 \\

 &\checkmark & & && 40.6 \\

 &\checkmark &  \checkmark&  && 40.0 \\
& &  & \checkmark & & 40.7 \\
& &  &  &\checkmark& 41.0 \\
& &  &  \checkmark&\checkmark& 40.6 \\
&\checkmark&  \checkmark&  \checkmark&\checkmark & 39.7\\
\bottomrule[1.4pt]
\end{tabular}
\end{table}

\subsubsection{Analysis of the Center Constraint} 
To analyze the effectiveness of the center constraints, we set $\alpha$ from 0.1 to 0.5 for experiments. As shown in Figure 5(a), the model gets improved results when $\alpha$ is set as 0.2, 0.3 and 0.4. When $\alpha$ is greater than 0.4 or less than 0.2, MR tends to go worse. That $\alpha$ is greater than 0.4 results in insufficient constraints. That $\alpha$ is less than 0.2 results in insufficient positive samples. 
\begin{figure}[!t]
\centering
\includegraphics[width=3.0in]{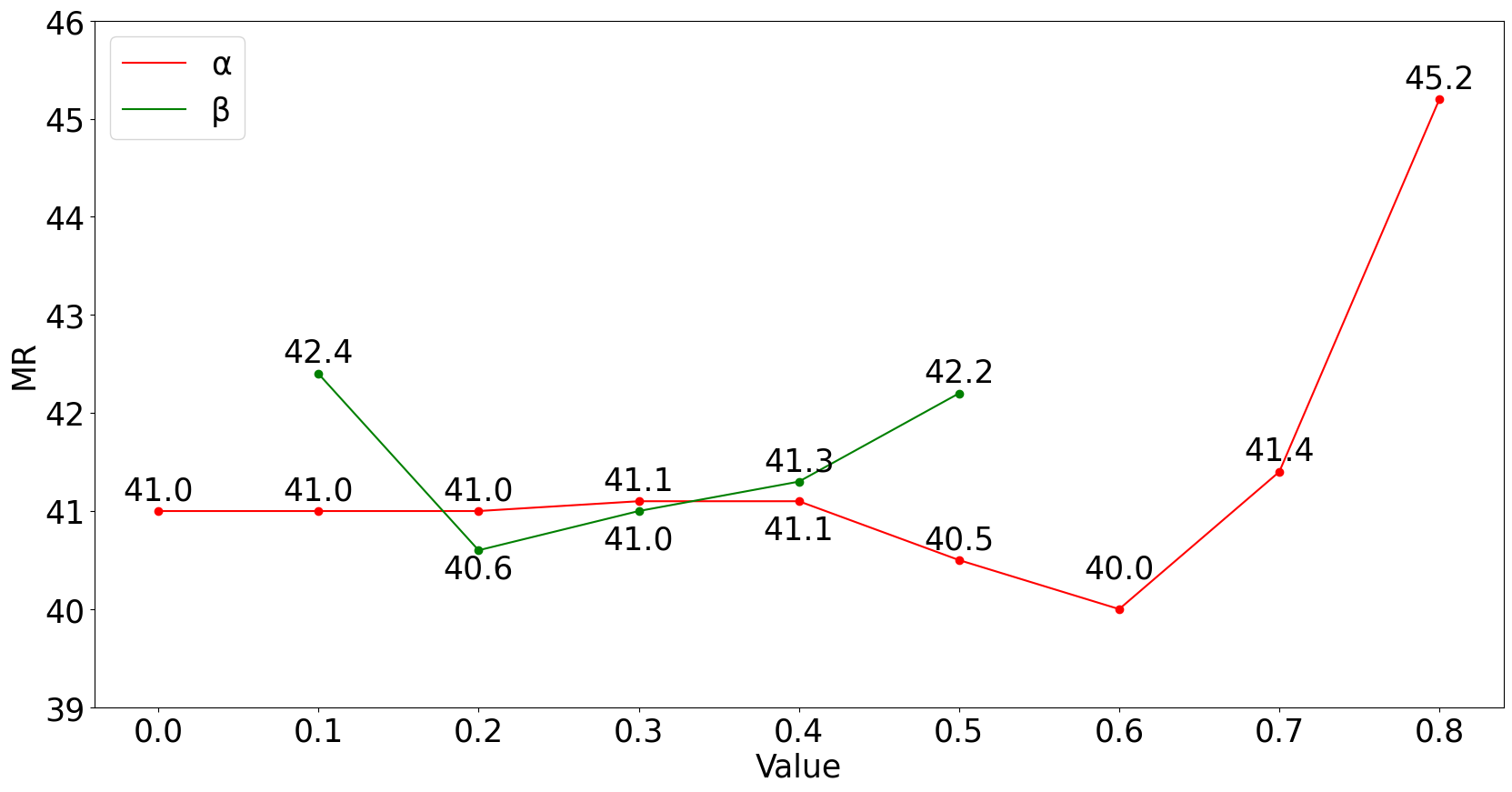}
\caption{Analysis of $\alpha$ and $\beta$.}
\label{fig3}
\end{figure}

\begin{figure}[!t]
\centering
\includegraphics[width=3.0in]{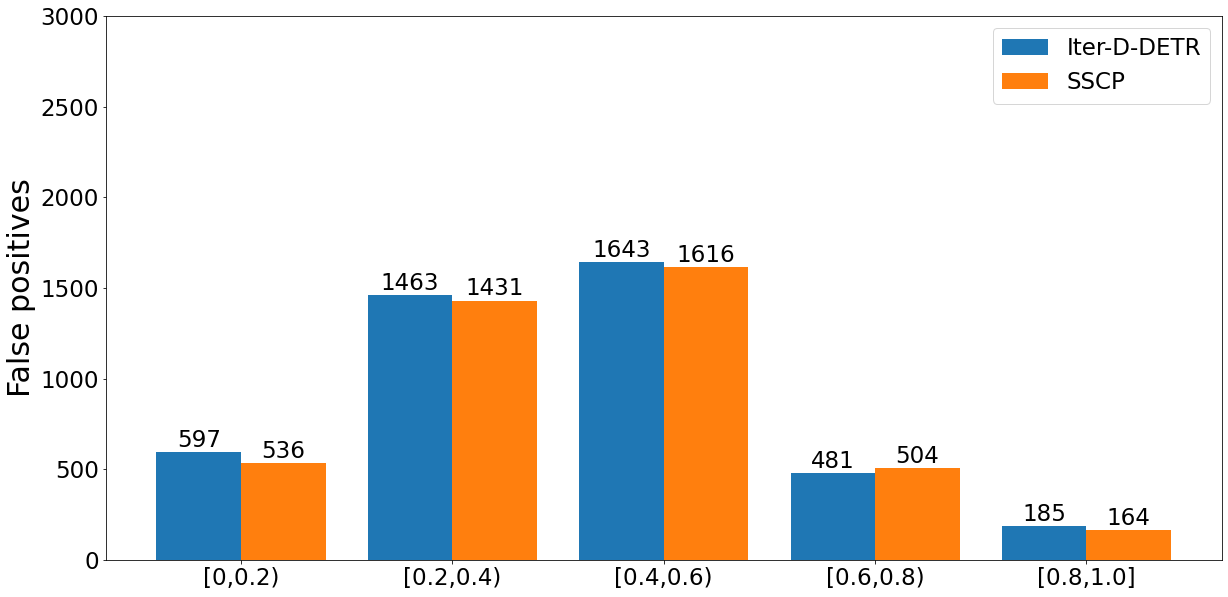}
\caption{Analysis of the improvement on MR.}
\label{fig3}
\end{figure}
\subsubsection{Analysis of the Position Constraint}  To analyze the effectiveness of the position constraint, we set $\alpha$ to 0.3 and set $\beta$ from 0 to 0.8 for experiments. As shown in Figure 5(b), $\beta$ from 0 to 0.4 does not show a difference in the results obviously. We think it is related to the evaluation metrics. MR uses 0.5 as the threshold to discriminate between false positives and true positives. The best result is achieved when the $\beta$ is set to 0.6. MR reaches 40.0\%. When the $\beta$ exceeded 0.6, performance shows a significant decline. We believe that it is due to insufficient positive samples for training.

\begin{figure*}[!t]
\centering
\includegraphics[width=7.0in]{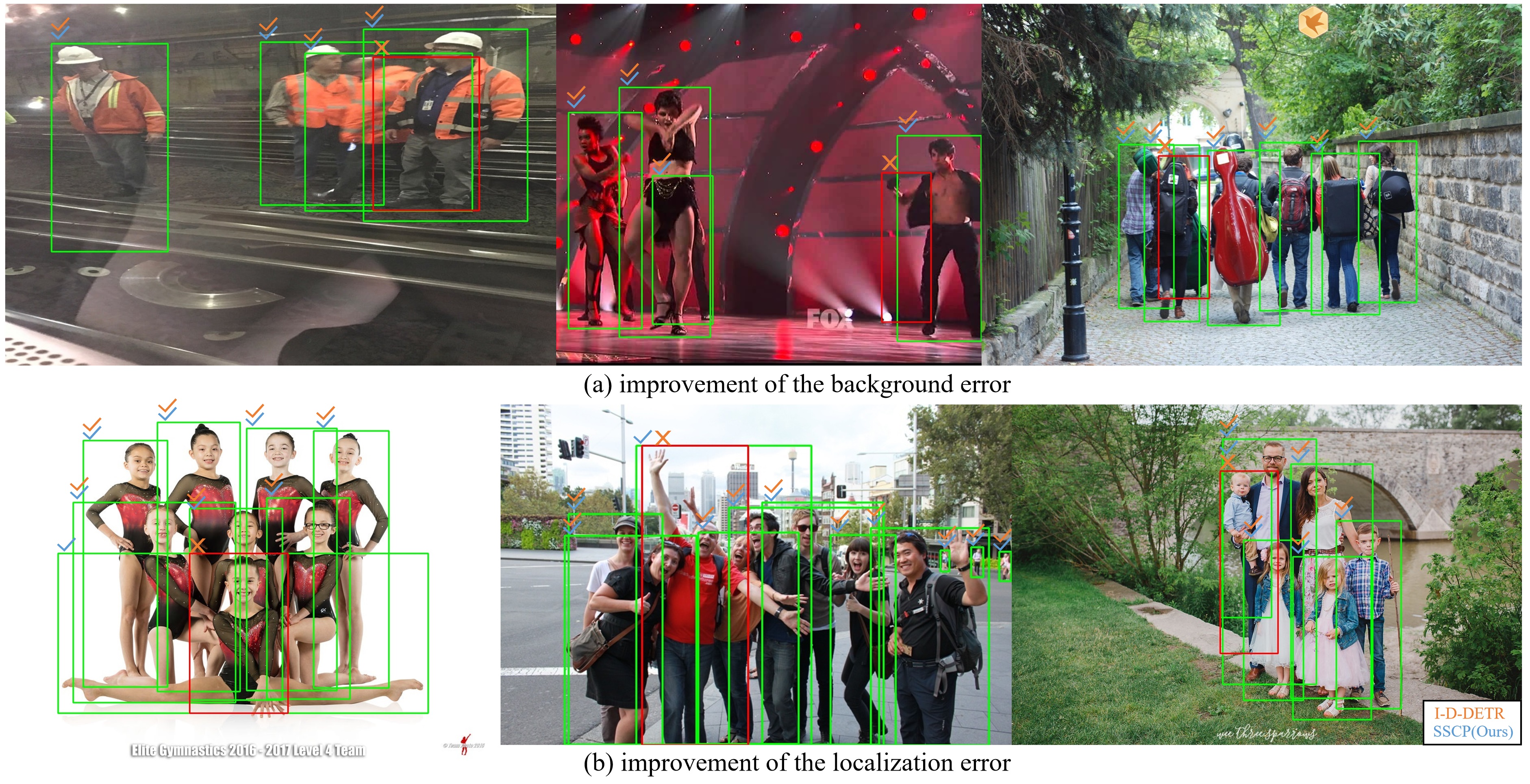}
\caption{The visualization results of our SSCP and our baseline Iter Deformable DETR. The box with the
check mark or cross represents the correct and false results. The green boxes are true positives produced by our SSCP. The red boxes are false positives produced by our baseline Iter Deformable DETR. In (a), Iter Deformable DETR detects the background as a pedestrian. In (b), IoU between the ground-truth and the prediction is less than 0.5, which causes the prediction to be a false positive. \label{tab:table}}
\label{fig6}
\end{figure*}

\subsubsection{Analysis of the combination of $\alpha$ and $\beta$ }  We set 4 sets of $\alpha$ and $\beta$, $\alpha=0.2,0.3$ and $\beta=0.5,0.6$, for experiments. When $\alpha=0.3$ and $\beta=0.6$, the model achieves the best performance 40.0\% MR on Crowdhuman dataset. When $\alpha=0.3$ and $\beta=0.5$, the model achieves a relatively poor performance, 40.5\% MR on Crowdhuman dataset. However, the performance is still obviously better than the baseline. Our approach is robust to the change of $\alpha$ and $\beta$, because these four sets of parameters do not cause significant performance changes. 

\subsubsection{Analysis of UAFL}  To analyze the effectiveness of UAFL, we separately apply soft label $y$ and the adaptive $\gamma$ for experiments, as shown in Table 3. Applying the soft label $y$ to the baseline improves MR by 1.2\%. Applying the adaptive $\gamma$ to the baseline improves MR by 0.9\%. $\beta$ of the adaptive $\gamma$ is set as the same as the best $\beta$ in CGLA, $\beta=0.6$. Without CGLA, UAFL improves MR by 1.3\%. Finally, with SSCP, the baseline is improved to 39.7\% MR.

\subsection{Analysis of the improvement on MR} 
Inspired by TIDE \cite{tide}, we design a similar experiment to illustrate our improvement on MR. First, we extracted the same number of predictions from SSCP and the baseline. The number depends on the smaller of the two models' valid predictions for MR. Secondly, the interval of the predicted boxes' IoU for the gound-truth is [0,1]. We divide the interval [0,1] into 5 intervals. Finally, we count false positives in each interval, as shown in Figure 6. Except for the interval [0.6,0.8), in the other intervals, our SSCP produces fewer false positives.
In terms of the total number of false positives, SSCP is still superior to the baseline. This is the main reason why MR is improved. In the interval [0.6,0.8), the extra false positives attribute to the missing annotation of Crowdhuman dataset. In fact, in the interval [0.6,0.8), some false positives are true positives.

%\par In conclusion, our SSCP has achieved state-of-the-art performance on both benchmarks especially in handling crowded pedestrians.

\subsection{Visualization}
\par In Figure 7, we can see the visualization of our SSCP and our baseline Iter Deformable DETR. The results are consistent with our expectations. Our method effectively reduces false positives in two aspects. The box with the
check mark or cross represents the correct and false results. In Figure 7(a), to show the results more clearly, we only show the true positives detected by the SSCP as common true positives which are the green boxes. In Figure 7(a), both SSCP and Iter Deformable DETR detect all pedestrians. However, Iter Deformable DETR redundantly detects the background as pedestrians which are the red boxes. In Figure 7(b), both SSCP and Iter Deformable DETR detect all pedestrians. However, the predictions of Iter Deformable DETR which are the red boxes are judged as negative samples due to localization problems. In evaluation, the IoU of a true positive is more than 0.5 with the corresponding ground-truth. Otherwise, it is a false positive even though it has detected the body of the pedestrian. In summary, our SSCP  effectively ameliorates the false positives problem of DETRs in crowded pedestrian detection.

\section{Conclusion}
In this paper, we first analyze the pipeline of DETRs and point out that selecting positive training samples which are not learnable is the key factor to making DETRs produce more false positives in crowded pedestrian detection. Then, we propose a simple but effective sample selection method SSCP to improve DETRs which consists of CGLA and UAFL. Assembled with our method, Iter Deformable DETR achieves state-of-the-art results on two challenging pedestrian detection benchmarks Crowdhuman and Citypersons without introducing any additional overhead.

\clearpage
\bibliographystyle{named}
\bibliography{ijcai23}

\end{document}